\documentclass[10pt,twocolumn,letterpaper]{article}

\usepackage{times}
\usepackage{epsfig}
\usepackage{graphicx}
\usepackage{amsmath}
\usepackage{amssymb}

\usepackage{pifont}
\newcommand{\cmark}{\textcolor{green}{\ding{51}}}%
\newcommand{\xmark}{\textcolor{red}{\ding{55}}}%

\usepackage[pagebackref=true,breaklinks=true,letterpaper=true,colorlinks,bookmarks=false]{hyperref}

\begin{document}

\title{Towards the extraction of robust sign embeddings for low resource sign language recognition}

\author{Mathieu De Coster\\
IDLab-AIRO -- Ghent University -- imec\\
{\tt\small mathieu.decoster@ugent.be}
\and
Ellen Rushe\\
SFI Lero \& Trinity College Dublin, Ireland\\
{\tt\small ellen.rushe@tcd.ie}
\and
Ruth Holmes\\
SFI Lero \& Trinity College Dublin, Ireland\\
{\tt\small holmesru@tcd.ie}
\and
Anthony Ventresque\\
SFI Lero \& Trinity College Dublin, Ireland\\
{\tt\small anthony.ventresque@tcd.ie}
\and
Joni Dambre\\
IDLab-AIRO -- Ghent University -- imec\\
{\tt\small joni.dambre@ugent.be}
}

\maketitle

\begin{abstract}
Isolated Sign Language Recognition (SLR) has mostly been applied on datasets containing signs executed slowly and clearly by a limited group of signers. In real-world scenarios, however, we are met with challenging visual conditions, coarticulated signing, small datasets, and the need for signer independent models. To tackle this difficult problem, we require a robust feature extractor to process the sign language videos. One could expect human pose estimators to be ideal candidates. However, due to a domain mismatch with their training sets and challenging poses in sign language, they lack robustness on sign language data and image-based models often still outperform keypoint-based models. Furthermore, whereas the common practice of transfer learning with image-based models yields even higher accuracy, keypoint-based models are typically trained from scratch on every SLR dataset. These factors limit their usefulness for SLR. From the existing literature, it is also not clear which, if any, pose estimator performs best for SLR. We compare the three most popular pose estimators for SLR: OpenPose, MMPose and MediaPipe. We show that through keypoint normalization, missing keypoint imputation, and learning a pose embedding, we can obtain significantly better results and enable transfer learning. We show that keypoint-based embeddings contain cross-lingual features: they can transfer between sign languages and achieve competitive performance even when fine-tuning only the classifier layer of an SLR model on a target sign language. We furthermore achieve better performance using fine-tuned transferred embeddings than models trained only on the target sign language. The embeddings can also be learned in a multilingual fashion. The application of these embeddings could prove particularly useful for low resource sign languages in the future. 
\end{abstract}

\section{Introduction}
Sign language recognition (SLR) has various applications, including automatic sign language corpus annotation
\cite{momeni2022automatic}, sign language information retrieval \cite{duarte2022sign} and Sign Language Translation (SLT)
\cite{camgoz2018neural}. Within these applications, the task of the SLR model is to extract syntactic and
semantic information from video data that can be used in the downstream task.
 In order to generalize to as many people and surroundings as possible, an SLR model should be invariant to the camera's position and intrinsics, to environments and lighting, and to the specific characteristics of the person that is signing.  These characteristics not only include age, gender, and ethnicity, but also clothing, accessories,  individual body morphology, signing speed, and any other features that do not contribute to the linguistic content.  Sign language datasets are typically limited in size, which means
that SLR models trained on these datasets are prone to exacerbating bias and not
having robust feature extractors.

In recent years, several open source human pose estimation tools have been released. Among the most popular are OpenPose \cite{cao2017realtime}, MMPose \cite{mmpose2020},
and BlazePose \cite{bazarevsky2020blazepose} (used in MediaPipe \cite{mediapipe}).
These tools extract landmarks, also known as keypoints, from image data: these are estimates of the 2D or 3D positions of joints in the human body and other important landmarks of the face and head.
These models tend to be trained on a large variety of images of human poses, so they are supposedly robust against the factors that may be confusing for SLR models \cite{moryossef2021evaluating}. Recent work has shown that models trained with
pose estimator features are more signer independent
sign language recognizers than image-based models,
especially in low resource scenarios \cite{holmes2022improving}. This is important as in out-in-the-wild applications, individuals will differ from those in the training set.

In the current scientific literature, keypoint-based SLR appears to be
limited by three factors. First, any errors made by the pose estimator will be carried over to any downstream SLR model. Second, pose estimators are often trained on general data (i.e., not on sign language data). This can result in sub-optimal performance when applied to sign language data that often involves fast and precise hand movements. For instance, MediaPipe Holistic may fail
to predict hand keypoints in certain cases of inter-manual or manual-facial interaction
\cite{moryossef2021evaluating}, and OpenPose may produce noisy keypoints \cite{de2020sign,orbay2020neural}.
A third factor is the way keypoints are used in SLR models. The predicted keypoint coordinates are often used as raw inputs to sequential models such as LSTMs or Transformers \cite{de2020sign,orbay2020neural}. However, the exact positions of keypoints are not as informative to SLR models as the \emph{relations} between keypoints.


These limitations can be tackled with a more tailored approach to pre-processing of raw keypoint predictions. For the first two,
we propose a novel post-processing method that normalizes the predicted keypoints and imputes any missing values. For the final limitation we introduce a ``pose embedding'' (SignPose2Vec), 
a non-linear transformation that can be transferred and fine-tuned to specific datasets. SignPose2Vec captures morphological properties of signs and is generic and language agnostic. By being based on keypoints, it is furthermore not tied to specific datasets and can be used as the starting
block of any keypoint-based SLR or SLT model.

The contributions of this paper are as follows. First, we compare three pose estimation tools for SLR, a comparison distinctly lacking in the SLR literature. Second, we show that transfer learning is possible with keypoint embeddings, even without fine-tuning. Transfer learning in SLR was previously only attempted with image-based models. Third, we present a method to train sign representations from multilingual data. Fourth, we show why performing isolated SLR on real-world data is more challenging than isolated SLR on individual productions of signs and how it can nonetheless be tackled.

\section{Related work}
\subsection{Human pose estimation for SLR}
Human pose estimation is the task of predicting the positions of specific points in the human body. These \emph{keypoints}
are often related to joints (e.g., the shoulders, elbows and wrists), but also include other pose landmarks such as
the positions of the eyes and ears. A multitude of human pose estimation models exist; here we give an overview
of toolkits that are commonly used in SLR research.

We compare two approaches towards pose estimation: top-down
and bottom-up. A top-down pose estimator first detects the individual(s)
in an image, and then detects and predicts keypoints for the constituent
body parts. A bottom-up pose estimator detects and predicts keypoints
for body parts and assigns them to individuals afterwards.

In 2017, the first open source real time full body multi person human pose estimation model
OpenPose was released \cite{cao2017realtime}.
It uses a bottom-up approach,
detecting body parts and aligning them to individuals using 
Part Affinity Fields (PAFs).
First, body part heatmaps are predicted along with the PAFs that indicate the direction from one body part
to another. A greedy algorithm is used to match body parts to the individuals in the image. Finally,
the keypoints are derived from the body part heatmaps.

MediaPipe Holistic is a top-down pose estimator that combines several neural networks into a unified pipeline
for body, face, and hand pose recognition. BlazePose \cite{bazarevsky2020blazepose} is used to detect and track the
pose throughout video data.
The pose keypoints are used to crop hand and face images that are passed on to specialized pipelines.
The hand model first detects the presence of the palm to refine the region of interest.
Then, coordinate regression is performed to obtain the hand keypoint coordinates.
Finally, all landmarks are merged into a single result.

MMPose is a software library that includes several computer vision
algorithms related to pose estimation \cite{mmpose2020}.
For example, it is often used as a top-down pose estimator. Faster R-CNN \cite{ren2015faster} detects
and crops the individual in the video, and HRNet \cite{sun2019deep} predicts keypoints. HRNet uses heatmap
regression to predict keypoint coordinates and confidence values for the full body.


Given a single (i.e., monocular) input image, OpenPose and HRNet predict 2D keypoint coordinates. Even though it is an
ill-posed problem, techniques exist to ``lift'' these 2D coordinates into 3D \cite{martinez2017simple}.
Alternative approaches directly predict depth from monocular data, as is the case in BlazePose which regresses keypoints from images \cite{bazarevsky2020blazepose} (as part of the MediaPipe software package).

OpenPose, MediaPipe, and MMPose have all been applied in domains such as action recognition \cite{shi2019skeleton,yan2020real}, gesture recognition \cite{qiao2017real,yadav2019real}, and SLR \cite{ko2018sign,de2019towards,li2020word,de2020sign,hruz2022one,jiang2021skeleton,konstantinidis2018deep,de2021isolated,moryossef2021evaluating}. To the best
of our knowledge, there exists no comparative analysis of the accuracy or run-time performance of these three toolkits on sign language data.

\subsection{Keypoint-based SLR}
SLR is the process of extracting syntactic and/or
semantic information from video data containing signing for use in downstream tasks.
\emph{Isolated} SLR is one way to perform this extraction. It is a video classification task: given a video clip of a single sign,
the goal is to predict which sign is performed.
Individual signs are distinct meaningful elements in sign language. They map to concepts such as nouns and verbs,
and in some cases to clauses \cite{sandler2001natural}.
Isolated SLR models can be used for the automatic annotation of sign language corpora \cite{de2019towards,mukushev2022towards,momeni2022automatic}. These models
can also be used in a transfer learning set-up, for example as feature extractors
for SLT models \cite{albanie2020bsl,shi2022open}.

SLR has evolved similarly to other subfields of computer vision, in that early works used handcrafted features \cite{zaki2011sign},
but in the past decade the field has moved towards automatic feature extraction with deep learning \cite{pigou2014sign}. Even more recently, SLR research has moved towards using only monocular RGB data,
instead of RGB-D data extracted with sensors such as the Microsoft Kinect. The results of an SLR challenge held at CVPR 2021 show that RGB models can achieve accuracy on par with RGB-D models \cite{sincan2021chalearn}.

These RGB videos are typically processed using Convolutional Neural Networks (CNNs). A first approach factorizes the spatial and temporal processing. De Coster et al. use a ResNet to
extract a feature vector for every frame in the video clip, and then process the sequence of feature vectors
with an LSTM or transformer \cite{de2020sign}. Li et al. use VGG16 as a feature extractor, and a GRU to process
the sequences \cite{li2020word}. In contrast, 3D CNNs allow for spatio-temporal processing. I3D is a popular 3D CNN architecture for SLR \cite{albanie2020bsl,sarhan2020transfer,varol2021read}.
Mino et al. have shown that both kinds of models (factorized and spatio-temporal) reach similar levels of accuracy on the same data \cite{mino2022effect}. Yet, the optimal model is of course dataset specific, especially in low resource settings. Note that factorized models allow processing variable length sequences without resorting to average- or max-pooling: this is especially useful because of the large differences in sample durations in real-world sign language data.

Due to the low resource nature of many SLR datasets, human pose estimators (as pre-trained models)
are a popular alternative for CNNs (trained on sign language data).
Li et al. use graph convolutional networks for SLR on the WLASL dataset \cite{li2020word}.
The state of the art pose-based SLR on the AUTSL dataset \cite{sincan2020autsl} is reached
by Vazquez et al., who use ``Multi-Scale Graph Convolutional Networks'' \cite{vazquez2021isolated}, and Jiang et al., who propose a similar ``Sign Language Graph Convolutional Network'' \cite{jiang2021skeleton}.
Ko et al.
use OpenPose keypoints in combination with a GRU network and SVM classifier on the KETI dataset \cite{ko2018sign}.
Moryossef et al. compare OpenPose and MediaPipe Holistic, and conclude that they perform similarly on the AUTSL dataset, but that both have failure cases that negatively impact the performance of their downstream SLR models \cite{moryossef2021evaluating}. De Coster et al. extract motion features from OpenPose keypoints to augment visual
representations, but do not evaluate the performance of these features on their own \cite{de2021isolated}.
Konstantinidis et al. use OpenPose keypoints as inputs to an SLR model
and compare the performance to image-based models on the LSA64 dataset \cite{konstantinidis2018deep}. De Coster et al. also perform such a comparison on the Corpus VGT \cite{van2015het}, and, like Konstantinidis et al., find that their image-based models outperform their pose based models \cite{de2020sign}.
This is somewhat unexpected, especially with smaller datasets, as extracting pose keypoints should alleviate some of the difficulty with learning signs from RGB data and remove much of the background noise associated with signing in-the-wild. Our paper shows why this is the case and how to make keypoint-based SLR models more powerful than image-based
SLR models.

\section{Keypoint-based SLR}
\subsection{Dataset}
\label{sec:dataset}
\begin{table}[]
\centering
\caption{Subset statistics for the VGT dataset.}
\label{tab:dataset}
\begin{tabular}{lll}
\hline
Subset     & Number of examples & Number of signers \\ \hline
Train      & 19267              & 88                \\
Validation & 2702               & 12                \\
Test       & 2998               & 11               \\ \hline
\end{tabular}
\end{table}

\begin{figure}[t]
\begin{center}
   \includegraphics[width=\linewidth]{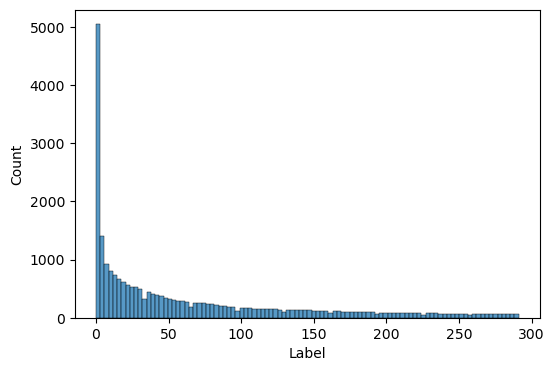}
\end{center}
   \caption{The class distribution of the entire VGT dataset shows a large class imbalance.}
\label{fig:class_distribution}
\end{figure}

We use the Flemish sign language corpus (Corpus VGT) \cite{van2015het} for our experiments.
This corpus is used for linguistics research and contains gloss (a transcription of a sign which can be used as a label in SLR) level annotations as well as, to a lesser extent,
translations into written Dutch.
We select all annotations for which the corresponding gloss has at least 20 occurrences, and only keep glosses
belonging to the established lexicon\footnote{Signs in the established lexicon have a known form, in contrast to productive signs which are created ``on-the-fly''.}. We obtain 26,639 examples belonging to 348 classes. We then perform
a stratified (on label) and grouped (on signer identity) split. Because the vocabulary
distributions vary between individual signers, we now have several glosses that do not occur in our training, validation, \emph{and} test set. We remove those glosses and are left with 24,968 examples belonging to 292 classes.
The resulting dataset has a long-tailed class distribution, that is shown in Figure \ref{fig:class_distribution}, similar to
the true sign distribution in the corpus \cite{bruynseraede2018lexicale}.
The head of the distribution consists of pointing signs, more specifically WG-1 (``I/me''), WG-2 (``you'') and WG-3 (``he/she/they/it'').
The dataset details are shown in Table \ref{tab:dataset}.

Despite every video corresponding to a single sign, the signs in this dataset are still influenced by coarticulation, which represents changes in the articulation of the sign due to the presence of other signs around it (e.g., the transitions between signs). This is an important difference with datasets such as AUTSL where signs are also \emph{performed}
in isolation. Coarticulation makes the isolated SLR problem more challenging.
Compared to working with datasets such as AUTSL or WLASL, the challenge is further compounded by the class imbalance. Moreover, in our dataset, there are different camera angles, whereas in many other SLR datasets, the camera is positioned in front of the signer.

We process every video with three keypoint extractors. For every frame in each video, we extract keypoints with
OpenPose, MMPose, and MediaPipe Holistic. We keep only the upper body and hand keypoints. In the case of MediaPipe
Holistic, we also remove the face mesh, which is high-dimensional and contains a large number of redundancies. In future work, keypoint
selection from the face mesh could allow us to model facial expressions and mouthings, which are disregarded in our
current pipeline. We obtain 54 2D keypoints for OpenPose, 53 2D keypoints for MMPose, and 67 3D keypoints for MediaPipe.

\subsection{Post-processing keypoints}
\label{sec:posepostproc}
We post-process the keypoint data before using them as inputs to the SLR model. For MediaPipe, this post-processing is applied to the 3D keypoints. When we choose to use 2D keypoints only, we drop the depth \emph{after} post-processing.
Our post-processing pipeline has two stages. We first perform missing keypoint imputation (only for MediaPipe and OpenPose) and then normalize the keypoints.

\paragraph{Imputation}
The imputation consists of linear interpolation, extrapolation, and zero-based imputation. For MediaPipe, we do this
separately for the right hand, the left hand, and the pose keypoints (for each of these subsets, MediaPipe will either return estimations for the entire subset, or return none at all, i.e., these are the smallest subsets of keypoints that MediaPipe
provides). For OpenPose, we perform imputation on a per-keypoint level, because individual keypoints can be missing.

Interpolation is applied to impute frames with missing data that are positioned
between two other frames with predicted keypoints.
Every missing frame is imputed by linearly
interpolating between the nearest non-missing previous and subsequent frames.

Extrapolation is applied for missing frames that occur in the beginning or at the end of a sequence. When no previous or subsequent frame exists, interpolation cannot be performed and we simply select the first
and last non-missing frame respectively and copy it over the missing frames.
There can also be samples where no predictions were available for a subset of the keypoints for the entire sequence. In these cases, we impute with zeros.

\begin{figure}[t]
\begin{center}
   \includegraphics[width=\linewidth]{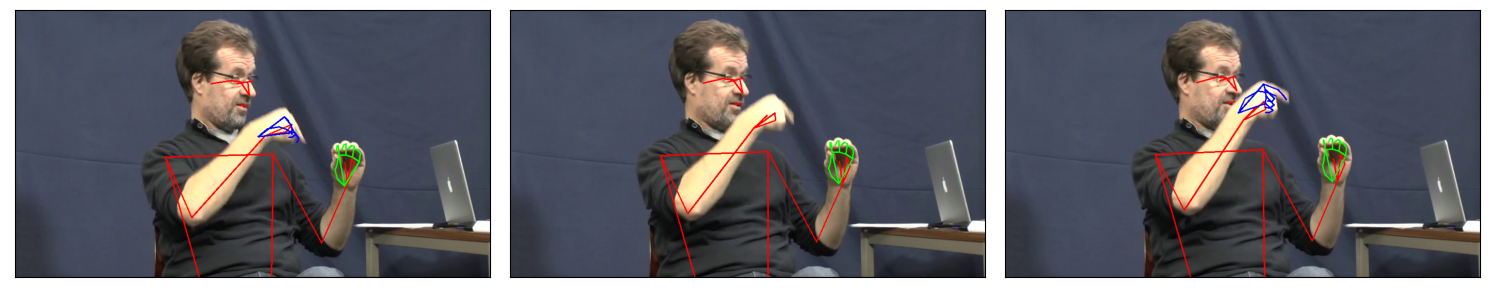}
\end{center}
   \caption{For cases such as these (the right hand is missing in the second frame but present in the first and third), linear interpolation can impute missing keypoints.}
\label{fig:interpolation_helps}
\end{figure}

Figure~\ref{fig:interpolation_helps} shows an example of a case where linear
interpolation can impute missing keypoints. Despite minor visual differences,
the pose estimator has failed to predict keypoints in the middle frame. We
correct this using the above approach.

\paragraph{Normalization}
For normalization, we account for differences in translation and scale of the keypoints. We first translate
the pose to be centered on the chest, and then re-scale all keypoints by dividing them by the Euclidean distance between the shoulders. We do the same for each hand, centering on the wrist and scaling by the Euclidean distance 
between the wrist and the knuckle of the middle finger.
This normalization step has a significant impact on the performance of keypoint-based models~\cite{bohavcek2022sign}.

\paragraph{Limitations}
\label{sec:posepostproclimits}
The selected pose estimation pipeline
only allows imputing keypoints that were not predicted. It is not possible to correct errors in predicted
keypoints. This is left for future research. To implement keypoint correction, confidence values (which are provided by OpenPose and MMPose but not by MediaPipe) are of great importance.

Furthermore, our imputation algorithm is limited. Extrapolation naively copies frames.
Linear interpolation fails to capture changes in acceleration and may impute intermediate hand shapes and body poses incorrectly for longer missing sub-sequences.
Interpolation is performed in the keypoint space (Cartesian coordinates). Linearly interpolating between these coordinates may result in physically impossible poses. Context-aware imputation should be investigated in future work.

Our normalization approach does not account for the rotation of the pose. Despite MediaPipe providing us with
3D keypoint data, giving us the ability to fully standardize the poses by rotating every pose to face the camera, we find that the depth predictions are not sufficiently robust to do this in a reliable way.

\subsection{Model architecture}
\label{sec:model}
\begin{figure}[t]
\begin{center}
   \includegraphics[width=0.5\linewidth]{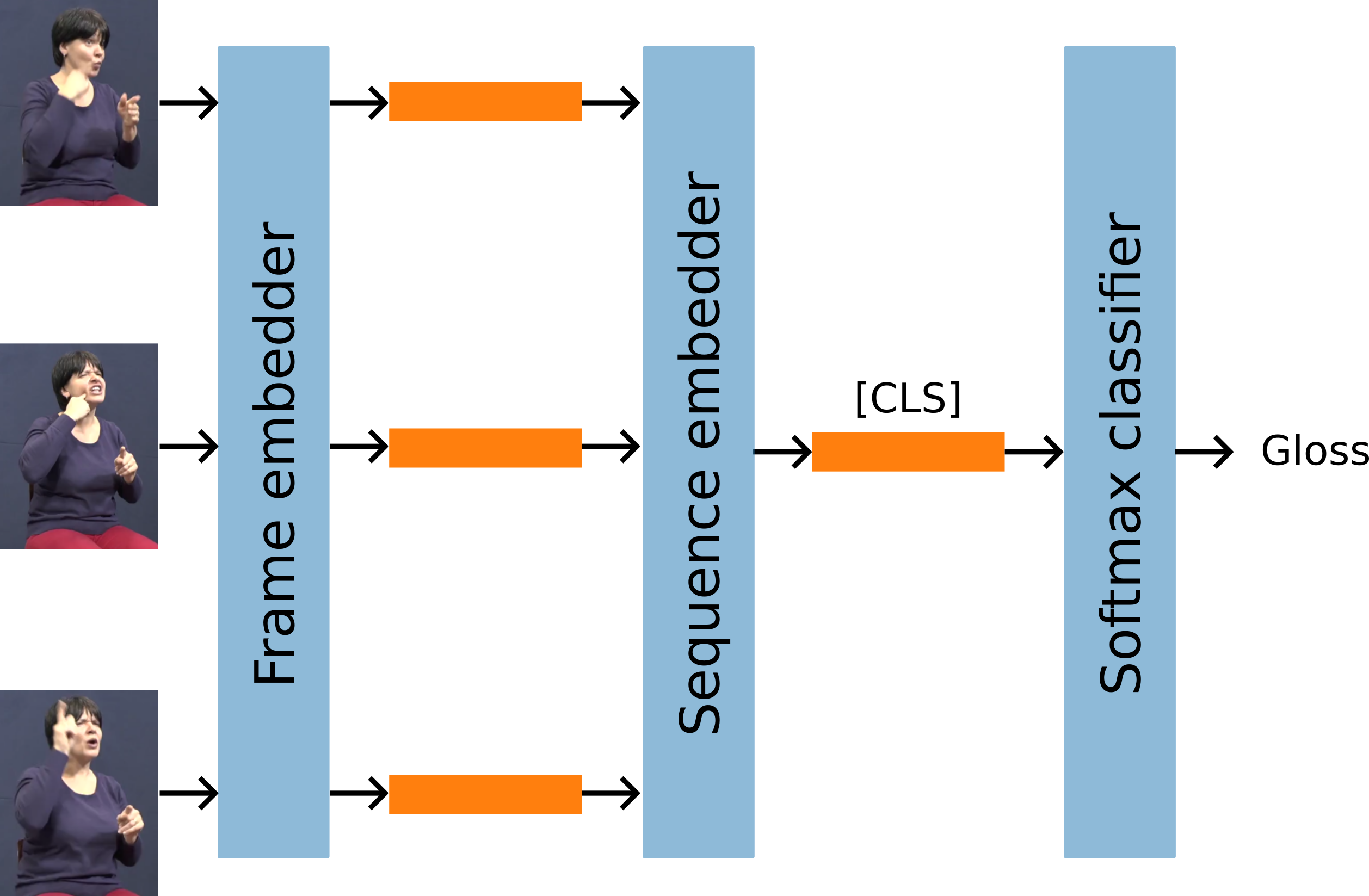}
\end{center}
   \caption{Generic three-stage model architecture for isolated SLR.}
\label{fig:architecture}
\end{figure}

Our models follow a three stage architecture that is illustrated in Figure \ref{fig:architecture}.
They consist of a frame embedder, a sequence embedder, and a softmax classifier.

The frame embedder varies based on the model being utilized.
For keypoint-based models, we employ a pose estimator followed by post-processing and finally by a dense pose embedding (see below).
In the case of image-based models (to which we compare our keypoint-based models), we utilize ResNet \cite{he2016deep} pre-trained on ImageNet \cite{deng2009imagenet}. Both the ResNet and pose embedding are applied to each individual frame in the sequence, resulting in a new sequence that is equally long as the original input.

The sequence embedder is a self-attention \cite{vaswani2017attention} network, which outperforms recurrent neural networks for isolated SLR \cite{de2020sign}. This self-attention network accommodates sequences of variable
length by zero padding shorter sequences in a batch and masking the padded frames in the attention computations.
We prepend to every sequence a learned vector, similar to the \verb+[CLS]+ token in BERT 
\cite{devlin2018bert}. The corresponding transformer output vector serves as the input to the classifier.

Unlike previous research using keypoint data as inputs for SLR models \cite{de2020sign,moryossef2021evaluating,konstantinidis2018deep,ko2018sign}, we propose learning an embedding (SignPose2Vec) on top
of the keypoints within the architecture described above. The purpose is to learn the non-linear relations between keypoints.
We will show that this embedding
can be transferred between languages and tasks, which is useful for pre-training purposes or for downstream translation models.
The pose embedding is implemented as a dense network of four blocks. The first three blocks consist of
a linear layer, normalized using layer normalization, followed by a ReLU activation,
and finally dropout ($p = 0.125$).
The final block has a hyperbolic tangent activation function instead of the ReLU and dropout. We apply L1 regularization to the input layer of the pose embedding ($\lambda = 0.002$) to focus on select keypoints.
We finally linearly transform the keypoints into the same dimension as the
embedding, and add the resulting vector element-wise to the embedding as
a residual connection.


\subsection{Experiments}
\paragraph{Monolingual SLR models}
We use a Video Transformer Network (VTN)---not to be confused with Video Vision Transformers \cite{arnab2021vivit}---as a baseline classifier. We use the architecture and hyperparameters proposed by De Coster et al. \cite{de2020sign}.
As feature extractor we use ResNet-34 pre-trained on ImageNet \cite{deng2009imagenet}, which produces a 512
dimensional frame embedding. The sequence of frame embeddings is processed using self-attention: in particular 4
transformer layers with 8 self-attention heads. We classify the output of the transformer (i.e., the \verb+[CLS]+ token
embedding) using a softmax classifier.

The Pose Transformer Network (PTN) is not directly adapted from De Coster et al. \cite{de2020sign}. Instead, we
prepend a pose embedding network to the self-attention network. This novel network is applied to every element of the sequence, producing a 128-dimensional embedding per frame.
Then, this embedding is processed similarly to the way it is in the VTN.
We compare MediaPipe Holistic (with 3D and 2D keypoints), OpenPose, and MMPose by using their post-processed
predicted keypoints as input to the PTN.

\paragraph{Variable length sequences}
For the VTN, we are limited to fixed length sequences due to the memory impact of processing many frames in parallel
with the ResNet. The PTN has a smaller memory footprint, allowing us to consider variable length
sequences. We zero-pad sequences that are shorter than the longest sequence in a batch and use a padding mask
to avoid that the model attends to padding.

\paragraph{Transfer learning}
We investigate the generalization ability of SignPose2Vec by first training a network on the NGT corpus, and then transferring the learned weights of the SLR model (except the classifier weights) to the VGT model. We also do this for the image-based VTN.
This dataset is constructed in a similar manner to the VGT dataset (Section \ref{sec:dataset}); it contains
68,854 samples for 458 classes in Dutch sign language (NGT). We only investigate transfer learning with the existing architecture and do not optimize the architecture for transfer learning: the performance reported here is a lower bound.

We perform three
experiments. First, we freeze the entire network except the classification layer, which we train on VGT.
Second, we do the same, but after the model has converged on VGT, we fine-tune the sequence embedding and classifier until early stopping.
In a third experiment, we fine-tune the entire model after convergence. This illustrates
the potential gains in accuracy with transfer learning.

\paragraph{Multilingual SLR models}
We consider training the model on multiple languages at the same time to learn more generic sign representations.
We train the model on the NGT dataset and the VGT dataset jointly.
Our approach is based on a paper on multilingual speech recognition by Toshiniwal et al.~\cite{toshniwal2018multilingual}. We compare two approaches to multilingual SLR in our first experiments.
In \emph{joint multilingual SLR}, we merge the vocabularies of NGT and VGT, and train a single classifier on the union of the vocabularies. This does not add additional hyperparameters, but signs that are shared or similar between languages will be mapped to different labels and this can introduce confusion.
In \emph{conditioned multilingual SLR} we pass the language ID to an embedding; the language embedding is an additional input to the classifier layer. Conditioning on the language identity should reduce confusions between signs from different languages that are similar. For the conditioned model, we use a 4-dimensional language ID embedding.


\section{Analysis}
\label{sec:poseanalysis}
\subsection{Comparison of pose estimators}
We compare MediaPipe Holistic, OpenPose, and MMPose on runtime performance and differences in outputs.

\begin{figure}[t]
\begin{center}
   \includegraphics[width=0.8\linewidth]{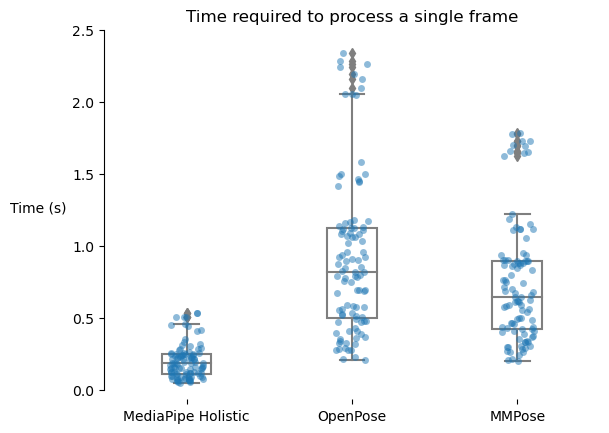}
\end{center}
   \caption{Runtime performance for the three pose estimators.}
\label{fig:runtime}
\end{figure}

We measure the runtime performance of MediaPipe, OpenPose and MMPose as follows:
we select a subset of 100 clips from our dataset. We then run all three pose estimators
on these clips and measure the execution time. We divide the execution time by the
number of frames in the video. We process a single video at a time to
simulate runtime performance at inference time.
These experiments were performed on an Intel Xeon Silver 4216 CPU (2.10GHz)
and an NVIDIA GeForce RTX 3090. In these experiments, MediaPipe runs on the CPU,
whereas OpenPose and MMPose leverage the GPU.
The results are shown in Figure~\ref{fig:runtime}. MediaPipe Holistic processes
4.8 frames per second (FPS), OpenPose 1.1 FPS and MMPose 1.3 FPS.

OpenPose and MMPose output confidence values per keypoint. MediaPipe Holistic does not.
Instead, if a hand could not be
detected or the pose estimation has failed, no values are returned for the entire
hand. The same is true for the body pose.
In the Corpus VGT (see Section \ref{sec:dataset}), we find that the body keypoints are missing in 0.005\% of the
frames, the left hand keypoints in 11\% of the frames and the right hand keypoints
in 8\% of the frames.

\begin{figure}[t]
\begin{center}
   \includegraphics[width=\linewidth]{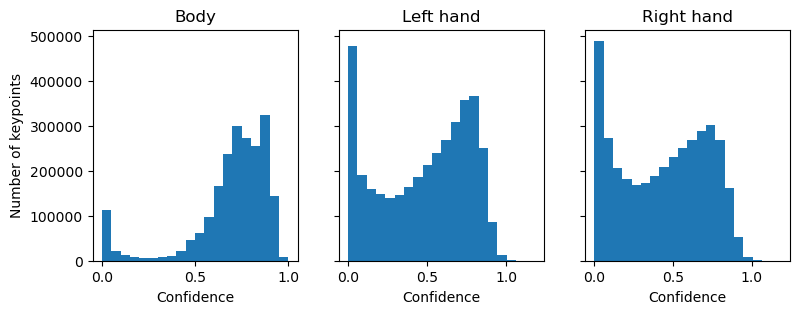}
\end{center}
   \caption{Distribution of confidence values returned by OpenPose.}
\label{fig:openpose_confidence}
\end{figure}

\begin{figure}[t]
\begin{center}
   \includegraphics[width=\linewidth]{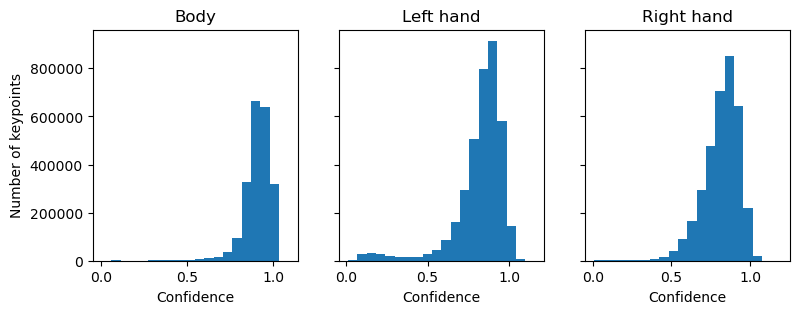}
\end{center}
   \caption{Distribution of confidence values returned by MMPose.}
\label{fig:mmpose_confidence}
\end{figure}

We observe in Figure~\ref{fig:openpose_confidence} that the hand keypoints predicted by OpenPose
have lower confidence values
than the body keypoints. This aligns with previous research that stated that
OpenPose body keypoints are more robust than hand keypoints \cite{de2021isolated}.
For the hands and the body, we observe peaks at 0 confidence.
These are keypoints that are mapped to the value $(0,0,0)$ as a fallback;
they are essentially ``missing values''.

Figure~\ref{fig:mmpose_confidence} shows that
MMPose is more confident in its predictions than OpenPose and there are no peaks at zero confidence.
Because in some cases there will be keypoints that MMPose should not be able to predict (e.g., when a hand is out of the frame), this suggests that MMPose may be more confident in its mistakes than OpenPose.
For the purpose of SLR, where precise keypoint predictions are important, we argue that having missing values that
can be imputed or corrected, is more useful than having
the pose estimator hallucinate keypoint coordinates.

\begin{figure}[t]
\begin{center}
   \includegraphics[width=\linewidth]{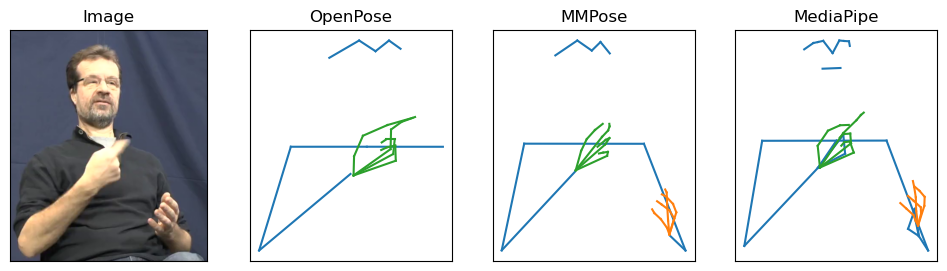}
\end{center}
   \caption{Keypoints predicted by OpenPose, MediaPipe, and HRNet. For OpenPose, the left elbow, wrist, and hand keypoints are missing.}
\label{fig:diffs_opmpmmp}
\end{figure}

The three pose estimators were trained on different data and have different architectures. Hence, their
predictions are different as well.
Figure~\ref{fig:diffs_opmpmmp} is illustrative of the difference between keypoints predicted by OpenPose, MediaPipe
Holistic, and MMPose. OpenPose fails to predict any keypoints for the left arm and hand. The hand keypoints in particular
show noise and variable predictions. Especially the right hand keypoints, which are the distinguishing
keypoints for this sign, are accurately predicted only by MediaPipe. 
MediaPipe Holistic has a dedicated hand pose estimation model, whereas OpenPose and MMPose use full-body models.
As a result, the hand keypoints are typically better in MediaPipe predictions (see the figure).

\begin{table}[]
\centering
\caption{Accuracy values for the models. The low training accuracy is due to heavy regularization to avoid overfitting.}
\label{tab:vtn_vs_ptn}
\begin{tabular}{lllll}
\hline
Model     & Pose estimator & Train & Validation & Test \\ \hline
VTN       & N/A & 52.42\% & 39.93\% & 38.93\% \\
PTN & MediaPipe (2D) & 76.61\% & \textbf{48.15\%} & 45.70\% \\
PTN & MediaPipe (3D) & 80.77\% & 45.93\% & 45.26\% \\
PTN & OpenPose & 77.90\% & 39.27\% & 35.86\% \\
PTN & MMPose & 62.72\% & 36.49\% & 36.00\% \\ \hline
\end{tabular}
\end{table}

These differences in keypoint predictions also appear in the downstream performance.
From Table~\ref{tab:vtn_vs_ptn}, it is clear that
MediaPipe outperforms OpenPose and MMPose. Interestingly, adding depth predictions from MediaPipe results
in worse performance. Visual analysis
informs us that these are not robust
and they
may simply introduce noise instead of
additional information. Considering that the test accuracies between MediaPipe with and without depth are more similar, it may be that the depth predictions are worse for certain individuals in the validation set.

\subsection{Variable length sequences}
We find that variable length sequences give significantly better performance than fixed length sequences
of window size 16 (the median number of frames in a sign) with stride 2.
The PTN trained on fixed length sequences achieves 40.56\% accuracy on the validation set,
whereas the PTN trained on variable length sequences achieves 48.15\%.
This is beneficial because it means that for inference, we can input entire sequences instead of using sliding windows.

\subsection{Pose post-processing}
\begin{table}[]
\centering
\caption{Ablation study results (validation accuracy).}
\label{tab:ablation}
\begin{tabular}{lllll}
\hline
Pose estimator & Norm. & Imputation & Accuracy \\ \hline
MediaPipe (2D) & \cmark & \cmark & 48.15\% \\
MediaPipe (2D) & \cmark & \xmark & 47.08\% \\
MediaPipe (2D) & \xmark & \xmark & 40.01\% \\
OpenPose & \cmark & \cmark & 39.23\% \\
OpenPose & \cmark & \xmark & 39.27\% \\
OpenPose & \xmark & \xmark & 35.53\% \\
MMPose & \cmark & N/A & 36.49\% \\
MMPose & \xmark & N/A & 21.47\% \\
\hline
\end{tabular}
\end{table}

We analyze the impact of our pose post-processing steps by performing an ablation study, removing imputation (for MediaPipe and OpenPose) and normalization from the pipeline. The results are shown in Table~\ref{tab:ablation}.
Normalization is clearly a crucial step in the pipeline. Missing data imputation is important for MediaPipe,
but less so for OpenPose. Whereas in OpenPose, individual keypoints can be missing, in MediaPipe it is always
an entire hand or more. Therefore, the impact of missing data is larger for MediaPipe.

\begin{figure*}[t]
\begin{center}
   \includegraphics[width=\textwidth]{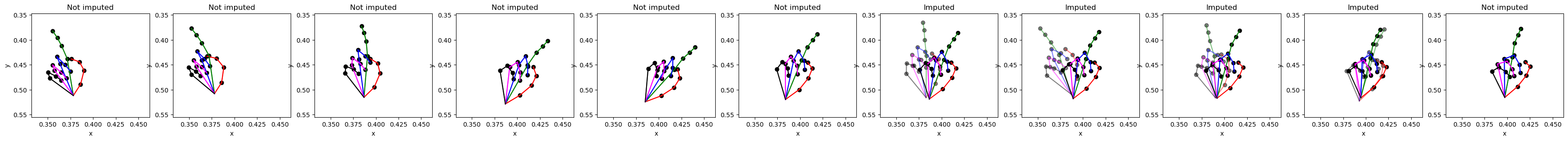}
\end{center}
   \caption{Example of a failure case for linear interpolation. The transparent hand is the ground truth,
   and the opaque hand is interpolated.}
\label{fig:interpolation_failure}
\end{figure*}

Interpolating shorter sequences works better than interpolating longer sequences. The longer the sequence, the higher the probability that the interpolation is incorrect. Figure~\ref{fig:interpolation_failure} shows an example: an intermediate hand pose was missed due to fast movements.
At the same time, when we only need to interpolate a short sequence, this missing data may not have as large of an impact. This is reflected in the small difference in score when using imputation and when not using imputation.

\subsection{Transfer learning with pose embeddings}
\label{sec:transfer}
\begin{table}[]
\centering
\caption{Transfer learning validation accuracy. $\Delta$ is the difference in accuracy between the model trained with transfer learning (NGT to VGT) and the model trained only on VGT data.
FT FE: Fine-tune frame embedding. FT SE: Fine-tune sequence embedding. Baselines are italicized.}
\label{tab:transfer}
\begin{tabular}{llllll}
\hline
Model     & FT FE    & FT SE & NGT     & VGT     & $\Delta$ \\ \hline
\em PTN & \em N/A & \em N/A & \em N/A & \em 48.15\% & \em N/A \\
PTN       & \xmark   & \xmark  & 44.19\% & 46.19\% & -1.96\% \\
PTN       & \xmark   & \cmark  & 44.19\% & 47.93\% & -0.22\% \\
PTN       & \cmark   & \cmark  & 44.19\% & \textbf{49.30\%} & +1.15\% \\
\hline
\em VTN & \em N/A & \em N/A & \em N/A & \em 39.93\% & \em N/A \\
VTN       & \xmark   & \xmark  & 38.81\% & 39.05\% & -0.88\% \\
VTN       & \xmark   & \cmark  & 38.81\% & 39.64\% & -0.29\% \\
VTN       & \cmark   & \cmark  & 38.81\% & 46.78\% & \textbf{+6.85\%} \\
\hline
\end{tabular}
\end{table}

Table~\ref{tab:transfer} shows the results of our transfer learning experiments. The frame and sequence embeddings can be transferred from NGT to VGT. Even when not fine-tuning
the embeddings, we achieve competitive (but slightly lower) accuracy values. This means that transfer learning to
languages without labeled data or with very limited amounts of labeled data is possible.
When fine-tuning the sequence embedding alone, and the sequence embedding and frame embedding together, we obtain better results.
The best results are obtained when fine-tuning the entire network after first fine-tuning the classifier layer.

We obtain a larger increase in accuracy when fine-tuning the VTN (compared to the PTN).
This is to be expected, because
the ResNet has more trainable parameters and thus can benefit significantly more from pre-training than our pose embedding.
Note that the VTN also needs to adapt to the different visual characteristics of the VGT dataset, unlike the PTN.
Nevertheless, this score (46.78\%) is still below that of a PTN trained only on VGT data (48.15\%).
For low resource sign languages, transfer learning of keypoint-based models specifically
will prove the most useful.

\subsection{Multilingual SLR models}
\begin{table}[]
\centering
\caption{Multilingual SLR accuracy.}
\label{tab:multilingual}
\begin{tabular}{lrrr}
\hline
Method & NGT accuracy & VGT accuracy \\
\hline
Joint       & 42.61\% & 38.27\% \\
Conditioned & 44.29\% & 49.52\% \\
\hline
\end{tabular}
\end{table}

The results of our multilingual learning are shown in Table~\ref{tab:multilingual}.
The joint multilingual model performs worse than the monolingual models (-1.58\% for NGT and -9.88\% for VGT).
However, the conditioned multilingual model outperforms the monolingual models (+0.19\% for NGT and +1.37\% for VGT). The difference reduces when fine-tuning the monolingual model (0.22\%).
Likely, there are similar signs in the NGT and VGT dataset which introduce confusions in the joint model. These
confusions can be eliminated by conditioning on the language embedding.

The multilingual models do not significantly outperform the monolingual models with transfer learning,
but training them is easier as it does not require a three step approach (pre-train, transfer,
fine-tune) and as only a single model needs to be trained instead of one for every considered combination of languages.

\section{Conclusion}
This paper tackles low resource spontaneous sign language recognition (SLR).
SLR models can be used for various tasks, such as automatic sign language corpus annotation, sign language information retrieval, and sign language translation. This paper investigates the extraction of robust
sign embeddings, a crucial step for these tasks.

In previous research, image-based models have typically outperformed keypoint-based models for SLR. However,
image-based models are prone to exacerbating bias, especially in low resource scenarios. We leverage human pose estimation tools as feature extractors, because these pose estimators have been trained on much larger and more varied
datasets than what is available for SLR, and they should be less biased as a result. Comparing the three most popular
pose estimators OpenPose, MMPose, and MediaPipe, we find that MediaPipe is the fastest and most robust.
Nevertheless, it still makes errors in its predictions (approximately 10\% of the frames have missing hand keypoints).
We perform post-processing on the keypoints in the form of normalization and missing value imputation and show the importance of these steps. By applying these steps, we outperform image-based models that were previously more powerful
than keypoint-based models for SLR.

This paper furthermore introduces a pose embedding that enables transfer learning for keypoint models across datasets and even sign languages. Transferring the pose embedding to another language can
even be done without fine-tuning with minimal loss of accuracy. When fine-tuning \emph{is} applied, we observe an increase
in accuracy for the downstream language.

Training, transferring and fine-tuning SLR models for different languages can be cumbersome. We propose to learn
language-conditioned multilingual SLR models. These models do not significantly outperform the monolingual models
when the latter are fine-tuned, but they only require a single training loop and result in a single model: this
saves on computational cost during training and reduces model storage requirements.

Future research still needs to tackle the large class imbalance in SLR datasets and handle
larger vocabularies. Even though we are able to drastically improve performance using
our pipeline, more research is still required into pose estimation specifically for sign language
data and how to optimally use the resulting keypoint data.

\section*{Acknowledgements}
The images of signers in Figures~\ref{fig:interpolation_helps}, \ref{fig:architecture}, and \ref{fig:diffs_opmpmmp} are reproduced from the Corpus VGT project \cite{van2015het} (CC BY-NC-SA).

Mathieu  De  Coster's  research  is  funded  by  the  Research Foundation Flanders (FWO Vlaanderen):  file number 77410. This work has been conducted within the SignON project. This project has received funding from the European Union's Horizon 2020 research and innovation programme under grant agreement No 101017255.

{\small
\bibliographystyle{ieee_fullname}
\bibliography{bibliography}
}

\end{document}